\documentclass[10pt,twocolumn,letterpaper]{article}

\usepackage{cvpr}              

\usepackage[accsupp]{axessibility}
\usepackage{graphicx}
\usepackage{amsmath}
\usepackage{amssymb}
\usepackage{booktabs}

\usepackage[pagebackref,breaklinks,colorlinks]{hyperref}

\usepackage{footnote}
\usepackage{amstext}
\usepackage{amsmath}
\usepackage{amssymb}
\usepackage{booktabs}
\usepackage{comment}
\usepackage{multirow}
\usepackage{adjustbox}
\usepackage{amsfonts}
\usepackage{siunitx}
\usepackage{color}
\usepackage{colortbl}  
\usepackage{xcolor}
\usepackage{array}   

\usepackage[capitalize]{cleveref}
\crefname{section}{Sec.}{Secs.}
\Crefname{section}{Section}{Sections}
\Crefname{table}{Table}{Tables}
\crefname{table}{Tab.}{Tabs.}

\def\cvprPaperID{1645} 
\def\confName{CVPR}
\def\confYear{2023}

\begin{document}

\title{MSeg3D: Multi-modal 3D Semantic Segmentation for Autonomous Driving}


\author{\stepcounter{footnote}
	Jiale Li$^{1}$
	\quad Hang Dai$^{2*}$ 
        \quad Hao Han$^{3}$
	\quad Yong Ding$^{3*}$ \\
	$^1$College of Information Science and Electronic Engineering, Zhejiang University \\
	$^2$School of Computing Science, University of Glasgow\\
	$^3$School of Micro-Nano Electronics, Zhejiang University \\
	$^*${\tt\small Corresponding authors\{Hang.Dai@glasgow.ac.uk, dingyong09@zju.edu.cn\}.}
	\vspace{-5mm}
}

\maketitle

\begin{abstract}
LiDAR and camera are two modalities available for 3D semantic segmentation in autonomous driving. The popular LiDAR-only methods severely suffer from inferior segmentation on small and distant objects due to insufficient laser points, while the robust multi-modal solution is under-explored, where we investigate three crucial inherent difficulties: modality heterogeneity, limited sensor field of view intersection, and multi-modal data augmentation.
We propose a multi-modal 3D semantic segmentation model (MSeg3D) with joint intra-modal feature extraction and inter-modal feature fusion to mitigate the modality heterogeneity. The multi-modal fusion in MSeg3D consists of geometry-based feature fusion GF-Phase, cross-modal feature completion, and semantic-based feature fusion SF-Phase on all visible points.
The multi-modal data augmentation is reinvigorated by applying asymmetric transformations on LiDAR point cloud and multi-camera images individually, which benefits the model training with diversified augmentation transformations.
MSeg3D achieves state-of-the-art results on nuScenes, Waymo, and SemanticKITTI datasets. Under the malfunctioning multi-camera input and the multi-frame point clouds input, MSeg3D still shows robustness and improves the LiDAR-only baseline.
Our code is publicly available at \url{https://github.com/jialeli1/lidarseg3d}.
\end{abstract}

\vspace{-4mm}
\section{Introduction}\label{sec:intro}
\vspace{-1mm}
Scene understanding for safe autonomous driving can be achieved through semantic segmentation using camera 2D images and LiDAR 3D point clouds, which densely classifies each smallest sensing unit of the modality.
The image-based 2D semantic segmentation has been developed with massive solid studies \cite{fcn_tpami2017,pspnet_cvpr2017,danet_cvpr2019,ocrnet_eccv2020}. The camera image has rich appearance information about the object but severely suffers from illumination, varying object scales, and indirect applications in the 3D world.
Another modality, LiDAR point cloud, drives 3D semantic segmentation with laser points \cite{nuscens_dataset,waymo_cvpr2020,semKITTI_dataset,PanopticNuscenes_dataset}. Unfortunately, irregular laser points are too sparse to capture the details of objects. The inaccurate segmentation appears especially on small and distant objects.
The other under-explored direction is using multi-modal data to increase both the robustness and accuracy in 3D semantic segmentation \cite{pmf_iccv2021}.

\begin{figure}[!t]
	\centering
	\includegraphics[width=0.9\linewidth]{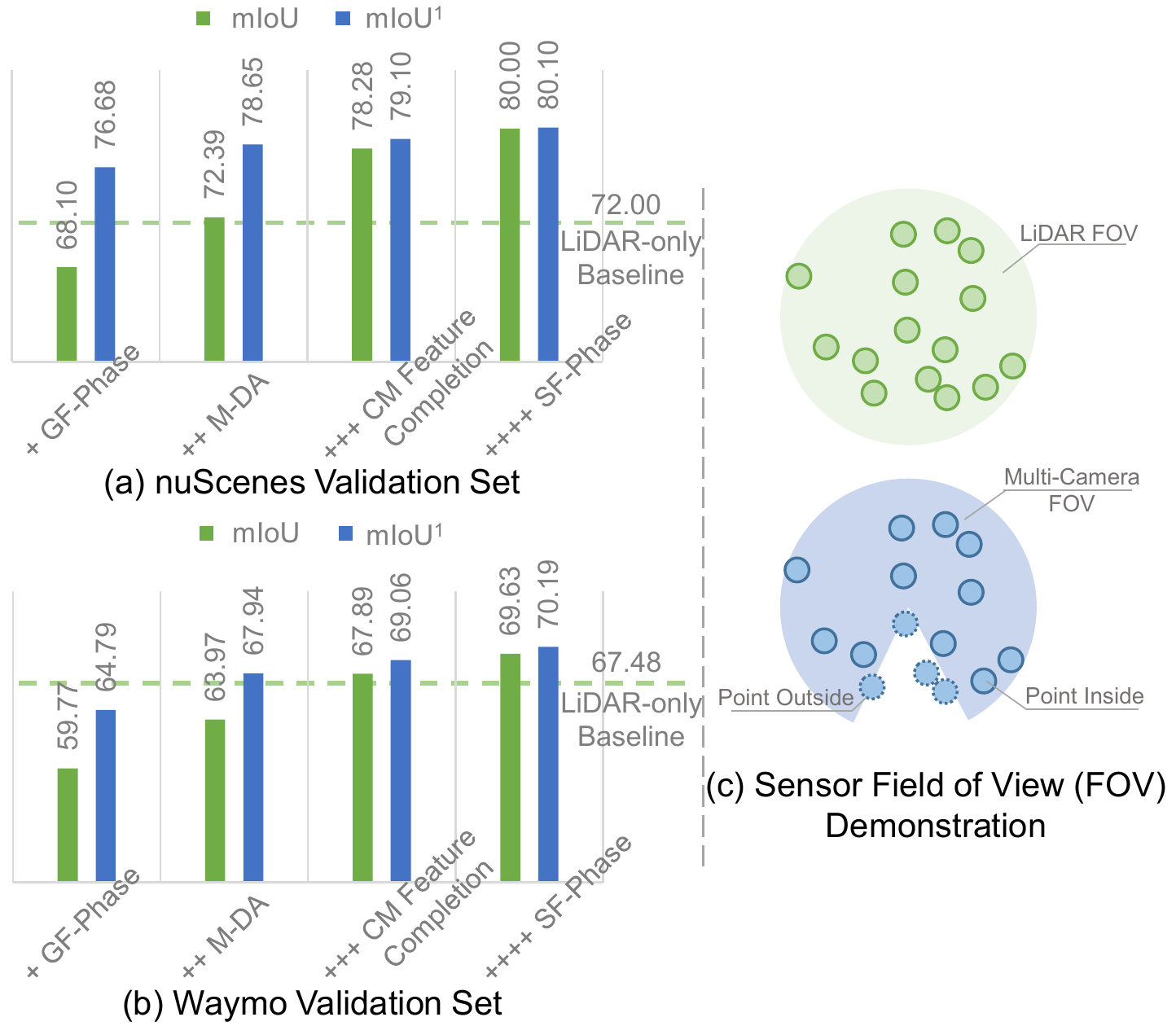}%
	\vspace{-3.5mm}
 	\caption{Performance (mIoU and mIoU$\textsuperscript{\ref{footnote:miou1}}$) evaluation on nuScenes \cite{nuscens_dataset} (a) and Waymo \cite{waymo_cvpr2020} (b) validation sets. All points are distinguished as points inside the sensor FOV intersection (points inside) and points outside the sensor FOV intersection (points outside). We cumulatively add (``+'') the proposed components.}
	\label{fig:miou_interior_vs_all}
	\vspace{-6mm}
\end{figure}

Despite the conceptual superiority, the development of multi-modal segmentation model is still nontrivial, lagging behind the single-modal methods \cite{nusclidarseg_leaderboard}.
We rationally attribute the difficulties to the three following aspects.
\textbf{i}) Heterogeneity between modalities.
Due to sparse points and dense pixels, point cloud feature extractors \cite{PintNet2_nips2017,submconv_cvpr2018} and image feature extractors \cite{resnet_cvpr2016,hrnet_tpami2021,segformer_nips2021} are developed distinctly. Separate intra-modal feature extractors are used to address the heterogeneity \cite{fuseseg_wacv2020,2d3dnet_3dv2021,RGBLIDAR_ITSC2019,pointpainting_cvpr2020,FusionPainting_ITSC2021}, but the lack of joint optimization leads to suboptimal features from irrelevant network parameters.
\textbf{ii}) Limited intersection on the field of view (FOV) between sensors.
Only the points falling into the intersected FOV are geometrically associated with multi-modal data, while simply considering the intersected multi-modal data is not sufficient to be practically applicable.
Performing fusion solely in the limited FOV intersection like \cite{pmf_iccv2021,fuseseg_wacv2020} results in unsatisfactory overall segmentation performance as shown in Fig.~\ref{fig:miou_interior_vs_all}.
\textbf{iii}) Multi-modal data augmentation. 
For example, PMF \cite{pmf_iccv2021} uses only several 2D augmentations for spatially aligned point cloud projection image and camera RGB image.
Under the constraint of modal alignment or 2D representation of point cloud, the multi-modal segmentation works \cite{pmf_iccv2021, fuseseg_wacv2020, RGBLIDAR_ITSC2019} discard many useful and critical point cloud augmentation transformations with sacrificed perception performance \cite{PIRCNN_AAAI2020, mvxnet_icra2019}.

Accordingly, we propose a top-performing multi-modal 3D semantic segmentation method termed MSeg3D, inherently motivated by addressing the aforementioned three technical difficulties. \textbf{i}) Unlike separately extracting modal features in existing methods \cite{2d3dnet_3dv2021,RGBLIDAR_ITSC2019,pointpainting_cvpr2020,FusionPainting_ITSC2021}, we jointly optimize intra-modal feature extraction and inter-modal feature fusion to drive maximum correlation and complementarity between heterogeneous modalities.
\textbf{ii}) To overcome the disregarded multi-modal fusion outside FOV intersection \cite{pmf_iccv2021, RGBLIDAR_ITSC2019}, we propose a cross-modal feature completion and a semantic-based feature fusion phase SF-Phase to collaborate with the geometry-based feature fusion phase GF-Phase. For points outside the FOV intersection, the former completes the missing camera features using predicted pseudo-camera features, under the explicit guidance of cross-modal supervision.
For all the points outside and inside the FOV intersection, the later SF-Phase leverages the multi-head attention \cite{attention_is_all_you_need_nips2017} to model the semantic relation between point and interested categories so that we can attentively fuse the semantic embeddings aggregated from all the visible fields to each point. 
\textbf{iii}) The challenging multi-modal data augmentation is reinvigorated by being decomposed as the asymmetric transformations in the LiDAR, camera worlds, and local cameras, which enables flexible permutation to enrich training samples.

As the proposed components accumulated in Fig.~\ref{fig:miou_interior_vs_all}, mIoU and mIoU\footnote{In the following text, mIoU$^{1}$ denotes the segmentation performance evaluated on only the points inside FOV intersection by excluding the points outside like PMF \cite{pmf_iccv2021} and other methods \cite{RGBLIDAR_ITSC2019, fuseseg_wacv2020}. \label{footnote:miou1}} are significantly increased 
while the gaps between mIoU and mIoU$\textsuperscript{\ref{footnote:miou1}}$ are gradually decreased.
Our contributions are four-fold:
\textbf{i})
We propose a multi-modal segmentation model MSeg3D with joint intra-modal feature extraction and inter-modal feature fusion, achieving state-of-the-art 3D segmentation performance on the competitive nuScenes \cite{nuscens_dataset}, Waymo \cite{waymo_cvpr2020}, and SemanticKITTI \cite{semKITTI_dataset} datasets for autonomous driving. The proposed framework won $2^{nd}$ place in the Waymo 3D semantic segmentation challenge at CVPR 2022.
\textbf{ii})
We propose a cross-modal feature completion and a semantic-based feature fusion phase. To our best knowledge, it is the first time to address the overlooked and inapplicable multi-modal fusion outside the sensor FOV intersection.
\textbf{iii})
By applying augmentation transformations asymmetrically on point cloud and images, the proposed asymmetrical multi-modal data augmentation significantly increases the diversity of multi-modal samples for training model with robust improvements.
\textbf{iv})
Extensive experimental analyses on the improvement and robustness of our method clearly investigate our designs.

\begin{figure*}[!t]
	\centering
	\includegraphics[width=1.0\linewidth]{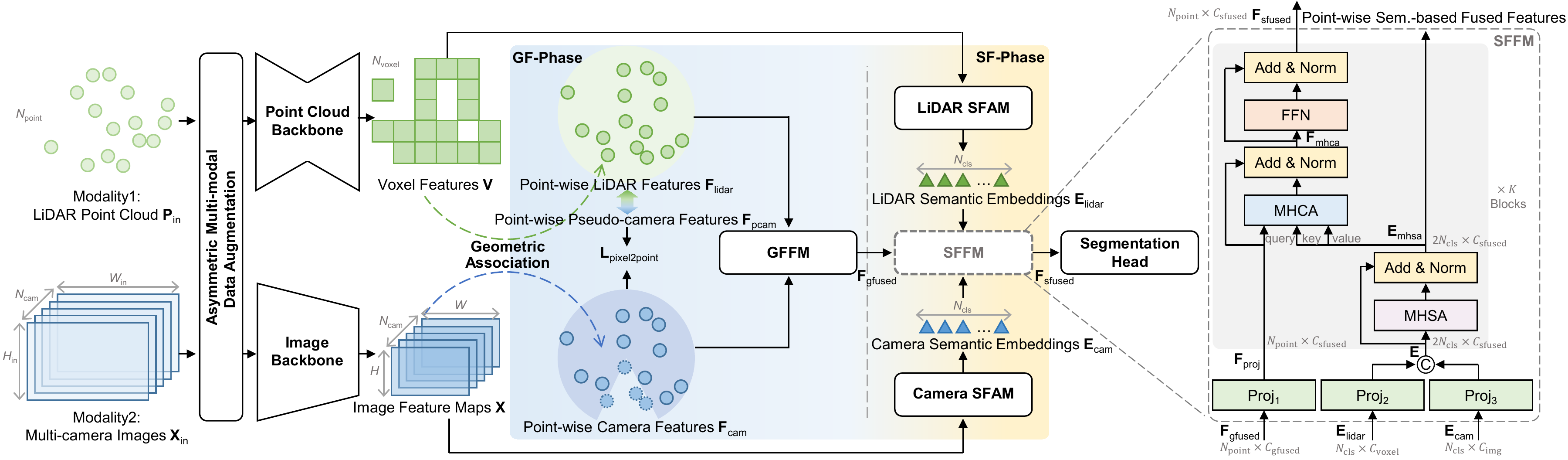}%
	\vspace{-2mm}
	\caption{Overview of our multi-modal 3D semantic segmentation model (MSeg3D). For multi-modal feature fusion, GF-Phase mainly includes the Geometry-based Feature Fusion Module (GFFM), while SF-Phase consists of LiDAR Semantic Feature Aggregation Module (SFAM), camera SFAM, and Semantic-based Feature Fusion Module (SFFM).}
	\label{fig:overview}
	\vspace{-4.5mm}
\end{figure*}

\section{Related Work}\label{sec:related_work}
\textbf{LiDAR-only 3D Semantic Segmentation} is promoted by SemanticKITTI \cite{semKITTI_dataset}, nuScenes \cite{nuscens_dataset}, and Waymo \cite{waymo_cvpr2020} datasets. The methods follow the U-Net \cite{unet_MICCAI2015} architecture, but progress from three point cloud representations.
\textbf{i}) \textbf{Point}.
The point methods derived from PointNet++ \cite{PintNet2_nips2017} cost heavy computation on sampling and gathering disordered neighbors, especially on a large-scale LiDAR point cloud. The major point methods \cite{PointASNL_cvpr2020, Randla-net_cvpr2020, kpconv_iccv2019, BAAFNet_cvpr2021} perform well on small synthetic point cloud \cite{ShapeNet_ACM2016} rather than sparse LiDAR point cloud.
\textbf{ii}) \textbf{2D images}.
PolarNet-series \cite{PolarNet_cvpr2020,PanopticPolarNet_cvpr2021} and others \cite{SqueezeSegV2_icra2019,squeezesegv3_eccv2020,salsanext_springer2020,range++_iros2019} project 3D point cloud as 2D images in bird’s-eye-view and range-view, achieving efficient LiDAR segmentation with 2D CNNs. But the 3D-to-2D projection damages 3D information and performance.
\textbf{iii}) \textbf{3D Voxel}.
The state-of-the-art SPVNAS \cite{spvnas_eccv2020}, Cylinder3D \cite{cylinder3d_cvpr2021} and SDSeg3D \cite{SDSeg3D_eccv2022} explicitly explore 3D structure information in different 3D coordinate systems, which  efficiently perform 3D sparse convolutions on non-empty voxels \cite{second_sensors_2018,MinkowskiNet_cvpr2019}. The point and image representations also potentially facilitate voxel features in RPVNet \cite{rpvnet_iccv2021} and AF2-S3Net \cite{AF2-S3Net_cvpr21}.
We also perform effective and efficient 3D voxel feature learning.

\textbf{Multi-modal 3D Semantic Segmentation} 
is investigated in \cite{pmf_iccv2021,fuseseg_wacv2020,RGBLIDAR_ITSC2019,2d3dnet_3dv2021} using LiDAR and camera with unsatisfactory performance.
The suboptimal feature sources of RGB value or separately learned CNN feature are warped into point cloud range image as additional input features \cite{RGBLIDAR_ITSC2019, fuseseg_wacv2020}. Recently, PMF \cite{pmf_iccv2021} projects point cloud onto the image plane for fusion with image data, which is obviously limited by camera FOV.
They discard the fusion and segmentation over points outside. Despite the improved segmentation performance within the FOV, another LiDAR model is required to take responsibility for the remaining area \cite{pmf_iccv2021, PointAugmenting_cvpr2021}. Such inequity and inconvenience have prevented researchers from focusing on multi-modal 3D semantic segmentation.
The image format of point cloud in the aforementioned works limits them to do only image augmentation such as 2D flipping rather than 3D point cloud augmentation. Instead, we point out the necessity of point cloud augmentation for our 3D segmentation network.
Unlike another multi-modal 3D object detector PointAugmenting \cite{PointAugmenting_cvpr2021} that preserves the point cloud augmentation while keeping the image unchanged, our method starts from both modalities with asymmetric transformations. 
Besides, 2D3DNet \cite{2d3dnet_3dv2021} aims to train a 3D model using reduced 3D annotations but more unlabeled data and additional manual 2D annotations to train a 2D model for obtaining 2D segmentation in advance. The 2D segmentation is also painted onto the points as the additional input of 3D model \cite{pointpainting_cvpr2020}. Such a phased training approach lacks joint 2D-3D optimization. Our motivation, technique, and results are distinguished from them.

\vspace{-1mm}
\section{Method}\label{sec:method}
\subsection{Preliminary}
\textbf{Problem Setup}.
Let $\{\mathbf{P}_{\text{in}}, \mathbf{X}_{\text{in}}\}$ be a multi-modal sample, where $\mathbf{P}_{\text{in}} \in \mathbb{R}^{N_{\text{point}} \times C_{\text{in}}}$ denotes LiDAR point cloud of $N_{\text{point}}$ points with $C_{\text{in}}$-dimensional input features (e.g., 3D coordinates and reflectance) and $\mathbf{X}_{\text{in}} \in \mathbb{R}^{N_{\text{cam}} \times 3 \times H_{\text{in}} \times W_{\text{in}}}$ denotes multi-camera RGB images of $N_{\text{cam}}$ cameras. Based on sensor calibration, the point with 3D coordinates $(x,y,z)$ can be projected onto the $c$-th local camera image with the pixel coordinates $(c,u,v)$. $N_{\text{cls}}$ semantic categories for each point in the 3D point cloud.

\textbf{LiDAR Feature Extraction}.
Given the input $\{\mathbf{P}_{\text{in}}, \mathbf{X}_{\text{in}}\}$ in Fig.~\ref{fig:overview}, we first conduct intra-modal feature extraction with parallel backbones for addressing the heterogeneity between point cloud and images. Due to the superiority of 3D voxel, we perform voxel-based LiDAR feature extraction by a general sparse 3D U-Net \cite{openpcdet2022,parta2_tpami2021} in Supplementary Material (SM). The input point cloud is first divided by a quantization step $d$, then grouped into non-empty voxels by averaging the point-wise initial features of the local points inside a voxel. These non-empty voxels $\mathbf{V}_{\text{in}} \in \mathbb{R}^{N_{\text{voxel}} \times C_{\text{in}}}$ constitute the sparse tensor input to point cloud backbone for learning the expressive voxel features $\mathbf{V} \in \mathbb{R}^{N_{\text{voxel}} \times C_{\text{voxel}}}$.

\textbf{Camera Feature Extraction}.
To enhance the LiDAR features with jointly-optimized camera features, we use a trainable image backbone (HRNet-w48 \cite{hrnet_tpami2021} by default) to non-linearly project $\mathbf{X}_{\text{in}}$ as $\mathbf{X} \in \mathbb{R}^{N_{\text{cam}} \times C_{\text{img}} \times H \times W}$, with down-sampled shape $H \times W$ and increased channels $C_{\text{img}}$.

The point cloud and image backbones can be flexibly selected from various mature networks. Moreover, we jointly optimize both backbones in the overall 3D segmentation model to learn relevant semantic feature representations from different modalities. The expressive voxel features $\mathbf{V}$ and image features $\mathbf{X}$ guarantee the inter-modal feature fusion, which consists of GF-Phase, cross-modal feature completion, and SF-Phase in the right half of Fig.~\ref{fig:overview}.

\subsection{GF-Phase: Geometry-based Feature Fusion}
The point-centric segmentation motivates us to warp voxel features $\mathbf{V}$ and image feature maps $\mathbf{X}$ to points in $\mathbf{P}_{\text{in}}$ as point-wise LiDAR features $\mathbf{F}_{\text{lidar}} \in \mathbb{R} ^{N_{\text{point}} \times C_{\text{voxel}}}$ and point-wise camera features $\mathbf{F}_{\text{cam}}  \in \mathbb{R} ^{N_{\text{point}} \times C_{\text{img}}} $ by using the geometric association of LiDAR and multi-camera.

\textbf{Learning Point Feature From Voxels and Pixels}.
We devoxelize voxel features $\mathbf{V}$ as point-wise LiDAR features $\textbf{F}_{\text{lidar}} = [f_{\text{lidar}, i}]^{i=N_{\text{point}}}_{i=1} \in \mathbb{R}^{N_{\text{point}}   \times C_{\text{voxel}} }$ for augmenting the point-wise fusion and segmentation. 
Given a point, we interpolate its point feature from the three nearest neighboring voxels \cite{FVTP_ACMMM2021}.
For the $i$-th point $(x_{i},y_{i},z_{i})$ with LiDAR feature $f_{\text{lidar},i}$ in $\textbf{F}_{\text{lidar}}$, we use the known point-wise image coordinates $(c_{i}, u_{i}, v_{i})$ to perform bilinear interpolation $\mathcal{I}$ on the image feature maps $\textbf{X}[c_{i}] \in \mathbb{R}^{ C_{\text{img}} \times H \times W } $ within the $c_{i}$-th local camera as Eq.~\ref{eq:bilinear_interpolate}. Zeros are temporarily padded to those points outside the camera FOV, which allows each point to be decorated with the $C_{\text{img}}$-dimensional feature $f_{\text{cam},i}$. To distinguish points outside, we set a binary mask $\mathbf{B}$ of $N_{\text{point}}$ elements with 0 for them and 1 otherwise.
\begin{align}
f_{\text{cam},i} &= \mathcal{I}( \textbf{X}[c_{i}], \frac{H}{H_{\text{in}}} u_{i}, \frac{W}{W_{\text{in}}} v_{i}), \label{eq:bilinear_interpolate}\\
\textbf{F}_{\text{cam}} &= [f_{\text{cam}, i}]^{i=N_{\text{point}}}_{i=1} \in \mathbb{R}^{N_{\text{point}}   \times C_{\text{img}} }. 
\end{align}

\textbf{Geometry-based Feature Fusion Module (GFFM)} 
first projects $f_{\text{lidar},i}$ and $f_{\text{cam},i}$ as $C_{\text{int}}$-dimensional by using fully connected layers $\mathcal{F}_{\text{lidar}}$ and $\mathcal{F}_{\text{cam}}$, and concatenates (``\copyright'') them together for subsequent learnable fusion using another MLP with $C_{\text{gfused}}$ output channels as $f_{\text{gfused}, i}$ $= MLP( \mathcal{F}_{\text{lidar}}( f_{\text{lidar},i} ) \copyright  \mathcal{F}_{\text{cam}}( f_{\text{cam},i} )  )$ and $\textbf{F}_{\text{gfused}}$ $= [f_{\text{gfused}, i}]^{i=N_{\text{point}}}_{i=1} \in \mathbb{R}^{ N_{\text{point}} \times C_{\text{gfused}} }$.
The GF-Phase combines the multi-modal features as $\textbf{F}_{\text{gfused}}$ mainly with geometry clues, so-called the geometry-based feature fusion phase.

\subsection{SF-Phase: Semantic-based Feature Fusion}\label{sec:sem_based_fusion}
Though the simple yet effective GF-Phase yields appreciable performance gains inside FOV intersection through unbiased considerations of both modalities, three drawbacks remain to be addressed.
\textbf{i})
The points outside still cannot retrieve their realistic camera features.
\textbf{ii})
The relative importance between modalities varies in different areas \cite{MVP_NIPS2021, BoundaryIoU_cvpr2021}, which is overlooked in GF-Phase.
\textbf{iii})
Explicit relational modeling for category-wise semantic representations \cite{isnet_iccv2021,mcibi_iccv2021} is not available.
As shown in Fig.~\ref{fig:overview}, we further propose the SF-Phase after GF-Phase. Beyond the geometric association in euclidean space, we aggregate LiDAR features and camera features into category-wise semantic embeddings as $\mathbf{E}_{\text{lidar}}$ and  $\mathbf{E}_{\text{cam}} $ for performing multi-modal fusion in a hidden semantic space.

\textbf{LiDAR Semantic Feature Aggregation Module (LiDAR SFAM)} starts from voxel features $\mathbf{V}$ and ends with aggregated $N_{\text{cls}}$ LiDAR semantic embeddings in $\mathbf{E}_{\text{lidar}}$.
For $N_{\text{cls}}$ categories, we assume $\mathbf{E}_{\text{lidar}} \in \mathbb{R}^{N_{\text{cls}} \times C_{\text{voxel}} } $ and a distribution matrix $\mathbf{D}_{\text{lidar}} \in (0,1)^{N_{\text{cls}} \times N_{\text{voxel}} }$. The $\mathbf{D}[j,i]$ closer to 1 indicates that the $i$-th voxel is more likely to belong to the $j$-th category than other voxels, so that it contributes more to describing this category in $\mathbf{E}_{\text{lidar}}[j]$. Otherwise, the $i$-th voxel should contributes less to $\mathbf{E}_{\text{lidar}}[j] $.
Now we can compute $\mathbf{E}_{\text{lidar}}$ by matrix product as $\mathbf{E}_{\text{lidar}} = \mathbf{D}_{\text{lidar}} \mathbf{V}$. Two more steps (Eqs.~\ref{eq:d_prime_lidar}$\sim$\ref{eq:d_transposed_lidar}) are required to compute $\mathbf{D}_{\text{lidar}}$. 
We first predict an intermediate segmentation $\mathbf{D}'_{\text{lidar}} \in \mathbb{R}^{N_{\text{voxel}} \times N_{\text{cls}}}$ on the voxel features $\mathbf{V}$ using an MLP-based auxiliary voxel segmentation head $\mathcal{H}_{\text{voxel}}$. We then perform a spatial softmax among voxels to normalize the values into $(0,1)$ for each category.
\begin{align}
\mathbf{D}'_{\text{lidar}} &= \mathcal{H}_{\text{voxel}}( \mathbf{V} ) \label{eq:d_prime_lidar},\\
\mathbf{D}^{T}_{\text{lidar}} &= [Softmax( \mathbf{D}'_{\text{lidar}}[:,j] )]^{j=N_{\text{cls}}}_{j=1} \label{eq:d_transposed_lidar}.
\end{align}
For more accurate $\mathbf{D}_{\text{lidar}}$, we explicitly guide $\mathbf{D}'_{\text{lidar}}$ by using the voxel segmentation supervision in Eq.~\ref{eq:loss_voxel}.

\textbf{Camera Semantic Feature Aggregation Module (Camera SFAM)} similarly aggregates $\mathbf{E}_{\text{cam}} \in \mathbb{R}^{N_{\text{cls}} \times C_{\text{img}} } $ from image feature maps $\mathbf{X} \in \mathbb{R}^{N_{\text{cam}} \times C_{\text{img}} \times H \times W}$ using another distribution matrix $\mathbf{D}_{\text{cam}} \in (0,1)^{N_{\text{cls}} \times N_{\text{pixel}} } $ and another auxiliary image segmentation head $\mathcal{H}_{\text{img}}$ implemented with the simple FCN head \cite{fcn_tpami2017}. The difference lies in that the camera SFAM is designed with all the $N_{\text{pixel}}$ pixels of image feature maps across $N_{\text{cam}}$ local cameras, where $N_{\text{pixel}}$ is $N_{\text{cam}} \times H \times W$.

The $\mathcal{H}_{\text{img}}$ first predicts the intermediate segmentation $\mathbf{D}'_{\text{img}} \in \mathbb{R}^{N_{\text{cam}} \times N_{\text{cls}} \times H \times W}$ by $\mathcal{H}_{\text{img}}( \mathbf{X} )$. Note that $\mathbf{D}'_{\text{img}}$ and $\mathbf{X}$ are then rearranged into $\mathbf{D}'_{\text{cam}} \in \mathbb{R}^{N_{\text{cls}} \times N_{\text{pixel}}}$ and $\mathbf{X'} \in \mathbb{R}^{N_{\text{pixel}} \times C_{\text{img}}}$ by transposing and reshaping operations. The final camera semantic embddings $\mathbf{E}_{\text{cam}}$ in Eq.~\ref{eq:e_cam} is the matrix product of $\mathbf{D}_{\text{cam}}$ and $\mathbf{X'}$, where the normalized $\mathbf{D}_{\text{cam}}$ is the result of applying spatial softmax on $\mathbf{D}'_{\text{cam}}$ as Eq.~\ref{eq:d_transposed_cam}.
\begin{align}
\mathbf{D}'_{\text{img}} &= \mathcal{H}_{\text{img}}( \mathbf{X} ),\label{eq:d_prime_img}\\
\mathbf{D}_{\text{cam}} &= [Softmax( \mathbf{D}'_{\text{cam}}[j,:] )]^{j=N_{\text{cls}}}_{j=1} \label{eq:d_transposed_cam},\\
\mathbf{E}_{\text{cam}} &= \mathbf{D}_{\text{cam}} \mathbf{X}' \label{eq:e_cam}.
\end{align}

Note that the image segmentation annotations are unavailable for guiding the $\mathbf{D}'_{\text{img}}$ in such 3D semantic segmentation datasets \cite{waymo_cvpr2020,nuscens_dataset}. Nevertheless, we also address this by the cross-modal semantic supervision scheme, which is detailed in Eq.~\ref{eq:y_img} and Eq.~\ref{eq:loss_img}.

\textbf{Semantic-based Feature Fusion Module (SFFM)}. 
As shown in the right dashed box in Fig.~\ref{fig:overview}, given the input of point-wise geometry-based fused features $\mathbf{F}_{\text{gfused}}$, category-wise semantic embeddings $\mathbf{E}_{\text{lidar}}$ and $\mathbf{E}_{\text{cam}}$, the SFFM first projects them into a $C_\text{sfused}$-dimensional space as $\mathbf{F}_{\text{proj}} = Proj_{1}(\mathbf{F}_{\text{gfused}}),\label{eq:proj1}$ and $\mathbf{E} = Proj_{2}(\mathbf{E}_{\text{lidar}}) \copyright  Proj_{3}(\mathbf{E}_{\text{cam}})$, respectively. Before stacking $K$ gray blocks in SFFM, we concatenate the projected multi-modal semantic embeddings together as $\mathbf{E} \in \mathbb{R}^{2N_{\text{cls}} \times C_{\text{sfused}} }$, which can be regarded as a dictionary that describes the typical characteristics for each category from the perspectives of LiDAR and cameras.

In each gray block, we model the semantic relations among all the $2N_{\text{cls}}$ category-wise semantic embeddings using the Multi-Head Self-Attention ($MHSA$) \cite{attention_is_all_you_need_nips2017} as Eq.~\ref{eq:e_mhsa}, where $Norm$ indicates the LayerNorm operation.
Let $C_{\text{shsa}}$ be $C_{\text{sfused}} / N_{H}$, the $MHSA$ can be decomposed into $N_{\text{H}}$ Single-Head Self-Attention ($SHSA$) operations with $\mathbf{E}_{h} \in \mathbb{R}^{2N_{\text{cls}}  \times C_{\text{shsa}} }$ as 
\begin{align}
    \mathbf{E}_{\text{mhsa}} &= Norm( \mathbf{E} + MHSA( \mathbf{E} ) ), \label{eq:e_mhsa} \\
    MHSA(\mathbf{E}) &= \left[  SHSA(  \mathbf{E}_{h} ) \right]_{h=1}^{h=N_{\text{H}}}, \label{eq:mhsa_to_shsa}\\
    SHSA(  \mathbf{E}_{h} ) &= Softmax( \frac{ \mathbf{Q}_{h} \mathbf{K}^{T}_{h} }{ \sqrt{ C_{\text{shsa}} }} )   \mathbf{V}_{h}. \label{eq:shsa}
\end{align}

Since $\mathbf{E}$ includes $\mathbf{E}_{\text{lidar}}$ and $\mathbf{E}_{\text{cam}}$, the attention matrix $Softmax( \frac{ \mathbf{Q}_{h} \mathbf{K}^{T}_{h} }{ \sqrt{ C_{\text{shsa}} } }) \in \mathbb{R}^{2N_{\text{cls}} \times 2N_{\text{cls}}}$ intuitively includes four parts of relation modeling on $\textbf{E}_{\text{lidar}} \rightarrow \textbf{E}_{\text{lidar}}$, $\textbf{E}_{\text{lidar}} \rightarrow \textbf{E}_{\text{cam}}$, $\textbf{E}_{\text{cam}} \rightarrow \textbf{E}_{\text{lidar}}$, and $\textbf{E}_{\text{cam}} \rightarrow \textbf{E}_{\text{cam}}$. 

After updating $\mathbf{E}$ as $\mathbf{E}_{\text{mhsa}}$, we perform the further fusion between the point-wise projected features $\mathbf{F}_{\text{proj}}$ and the semantic embeddings $\mathbf{E}_{\text{mhsa}}$, using the Multi-Head Cross-Attention ($MHCA$) for semantic relation modeling in Eq~\ref{eq:f_mhca} and the Feed-Forward Network ($FFN$) for feature embedding in Eq.~\ref{eq:f_ffn}, respectively.
\begin{align}
    \mathbf{F}_{\text{sfused}} &= Norm( \mathbf{F}_{\text{mhca}} + FFN( \mathbf{F}_{\text{mhca}} ) ), \label{eq:f_ffn} \\
    \mathbf{F}_{\text{mhca}} &= Norm( \mathbf{F}_{\text{proj}} + MHCA(\mathbf{F}_{\text{proj}}, \mathbf{E}_{\text{mhsa}}, \mathbf{E}_{\text{mhsa}})    ). \label{eq:f_mhca}
\end{align}

Unlike Eq.~\ref{eq:e_mhsa}, the $MHCA$ in Eq.~\ref{eq:f_mhca} sets the query $\mathbf{Q}$ from point-wise feature $\mathbf{F}_{\text{proj}}$, and sets the key $\mathbf{K}$ and value $\mathbf{V}$ from semantic embeddings $\mathbf{E}_{\text{mhsa}}$. Thus, an attention matrix with shape $N_{\text{point}} \times 2N_{\text{cls}}$ is computed to attentively aggregate the more important semantic embeddings respective to individual points, beyond the paired geometric association in GF-Phase.
Following the common usages of multi-head attention \cite{detr_eccv2020}, $K$ (i.e., 6) gray blocks are stacked.

\textbf{Discussion on SFFM}.
As a useful and general technique \cite{segformer_nips2021, pointformer_cvpr2021,detr_eccv2020}, multi-head attention \cite{attention_is_all_you_need_nips2017}, although not proposed in this paper, is effectively tailored to multi-modal feature fusion with different motivations and designs by us.
\textbf{i})
In Eq.~\ref{eq:f_mhca}, the $MHCA$ enables the point-wise feature to attend to multi-modal semantic embeddings, so that both inside and points outside can be consistently supported by expressive multi-modal semantic embeddings. 
\textbf{ii})
The $MHCA$ attention matrix computes the relative importance of two modalities to each point, improving the unbiased considerations of modalities in GF-Phase. 
\textbf{iii})
The $MHSA$ in Eq.~\ref{eq:e_mhsa} explicitly models the category-wise intra-modal and inter-modal semantic relations, deriving the commonality learning.
\textbf{iv})
The LiDAR SFAM and Camera SFAM aggregate the long sequences of voxel features $\mathbf{V}$ and image features $\mathbf{X'}$ into the short sequences of $\mathbf{E}_\text{lidar}$ and $\mathbf{E}_\text{cam}$ with $N_{\text{voxel}} / N_{\text{cls}}$ and $N_{\text{pixel}} / N_{\text{cls}}$ times, which enables efficient computation of the multi-head attention.

\subsection{Cross-modal Feature Completion}
As shown in Fig.~\ref{fig:overview}, a cross-modal feature completion module with pixel-to-point loss $\mathbf{L}_{\text{pixel2point}}$ (Eq.~\ref{eq:loss_pixel2point}) is set, 
where the point-wise pseudo-camera feature $\mathbf{F}_{\text{pcam}}$ is mapped from the point-wise LiDAR features $\mathbf{F}_{\text{lidar}}$ by another MLP-based $\mathcal{H}_{\text{pcam}}$ as $\mathcal{H}_{\text{pcam}}( \mathbf{F}_{\text{lidar}} )$. In practice, we compute the mean square error loss $\mathcal{L}_{\text{mse}}$ between the $\mathbf{F}_{\text{pcam}}$ and $\mathbf{F}_{\text{cam}}$ of the points inside for learning from the correctly paired cross-modal features relationship. 
\begin{align}
    \mathbf{L}_{\text{pixel2point}} &= \mathcal{L}_{\text{mse}} (\mathbf{B} {\mathbf{F}}_{\text{pcam}},  \mathbf{B}  Detach( \mathbf{F}_{\text{cam}}  )) \label{eq:loss_pixel2point}.\\
    \mathbf{F}_{\text{cam}}[i,:] &= \mathbf{F}_{\text{pcam}}[i,:] \mid _{i \in \{j \mid \mathbf{B}[j]=0 \} } \label{eq:mix_pcam_cam}.
\end{align}
Note that the points outside are ignored by the binary mask $\mathbf{B}$, which is mentioned in GF-Phase. The gradients of $\mathbf{F}_{\text{cam}}$ are also detached for optimizing the learning of $\mathbf{F}_{\text{pcam}}$.

We employ such a cross-modal feature completion module due to two motivations:
\textbf{i})
Optimizing $\mathbf{L}_{\text{pixel2point}}$ forces LiDAR features $\mathbf{F}_{\text{lidar}}$ to imitate camera features $\mathbf{F}_{\text{cam}}$ with $\mathcal{H}_{\text{pcam}}$, then the learned pseudo-image features $\mathbf{F}_{\text{pcam}}$ can be switched to replace the padded zeros in $\mathbf{F}_{\text{cam}}$ as Eq.~\ref{eq:mix_pcam_cam} in the inference stage, which serves as the cross-modal feature completion to enhance the feature learning. Thus, we further reduce feature gaps between points outside and inside.
\textbf{ii})
$\mathcal{H}_{\text{pcam}}$ transfers rich appearance priors from the camera branch to the LiDAR branch with an effective consistency constraint for enhancing the intra-modal feature learning in the training stage.

\subsection{Cross-modal Semantic Supervision}\label{sec:cross_modal_supvervision}

\textbf{Point Supervision}.
Let $\mathbf{Y}$ of $N_{\text{point}}$ elements be the point-wise 3D semantic segmentation labels. The value of $\mathbf{Y}[i]$ is in $[0, N_{\text{cls}}-1]$, where 0 denotes the ignored category.
An MLP based point segmentation head $\mathcal{H}_{\text{point}}$ is built on $\mathbf{F}_{\text{sfused}}$ for the 3D segmentation prediction $\hat{\mathbf{Y}} = \mathcal{H}_{\text{point}}( \mathbf{F}_{\text{sfused}} )$. Following \cite{CENet_icme2022,cylinder3d_cvpr2021,SDSeg3D_eccv2022}, we adopt point loss $\mathbf{L}_{\text{point}}$ as a combination of cross-entropy loss $\mathcal{L}_{\text{ce}}$ and lovasz-softmax loss $\mathcal{L}_{\text{lovasz}}$ \cite{lovasz_loss_cvpr2018} as $\mathcal{L}_{\text{ce}}(\hat{\mathbf{Y}}, \mathbf{Y}) + \mathcal{L}_{\text{lovasz}}(\hat{\mathbf{Y}}, \mathbf{Y}) $.

\textbf{Point-to-voxel Supervision}.\label{sec:point2voxel_loss}
For guiding the $\mathbf{D}'_{\text{lidar}}$ in Eq.~\ref{eq:d_prime_lidar}, the voxel loss $\mathbf{L}_{\text{p2v}}$ is defined as: 
\begin{equation}
\begin{aligned}
    \mathbf{L}_{\text{p2v}} = \mathcal{L}_{\text{ce}}(\mathbf{D}'_{\text{lidar}}, \mathbf{Y}_{\text{p2v}}) + \mathcal{L}_{\text{lovasz}}(\mathbf{D}'_{\text{lidar}}, \mathbf{Y}_{\text{p2v}}),\label{eq:loss_voxel}
\end{aligned}
\end{equation}
where the voxel labels $\mathbf{Y}_{\text{p2v}}$ of $N_{\text{voxel}}$ elements can be determined from the labels of points in the voxel.
To avoid ambiguity in $\mathbf{L}_{\text{p2v}}$, the label of the voxel containing points of multiple categories is set to zero as the ignored category.

\textbf{Point-to-pixel Supervision}.\label{sec:point2pixel_loss}
The problem of how to use only point labels to correctly guide $\mathbf{D}'_{\text{img}}$ (Eq.~\ref{eq:d_prime_img}) is still unsolved.
For $\mathbf{D}'_{\text{img}}$ of each training sample, we initialize a $N_{\text{cam}} \times H \times W$ map $\mathbf{Y}_{\text{point2pixel}}$ of all zeros as the image segmentation label. We then project each point onto the down-sampled image planes and retrieve the point label $\mathbf{Y}[i]$ for the corresponding nearest ($\langle\cdot\rangle$) pixel as Eq.~\ref{eq:y_img}. Although the generated image label $\mathbf{Y}_{\text{point2pixel}}$ is sparse, it provides sufficient supervision.
In SM, Fig.~\textcolor{red}{S2} and Fig.~\textcolor{red}{S3} provide visualizations of $\mathbf{D}'_{\text{img}}$ to show that these sparse point-to-pixel supervisions can propagate to other regions with similar appearance, which guides $\mathbf{D}'_{\text{img}}$ correctly. Since $\mathbf{Y}_{\text{point2pixel}}$ is sparse, we only compute the basic cross-entropy loss $\mathcal{L}_{\text{ce}}$ in Eq.~\ref{eq:loss_img}.
\begin{align}
    &\mathbf{Y}_{\text{point2pixel}}[  c_{i}, \langle \frac{H}{H_{\text{in}}} u_{i} \rangle, \langle \frac{W}{W_{\text{in}}} u_{i} \rangle] = \mathbf{Y}[i], \label{eq:y_img}\\
    &\mathbf{L}_{\text{point2pixel}} = \mathcal{L}_{\text{ce}}(\mathbf{D}'_{\text{img}}, \mathbf{Y}_{\text{point2pixel}}). \label{eq:loss_img}
\end{align}

\textbf{Total Loss Function}.
To jointly optimize all the modules in the proposed segmentation model, the total loss combines loss terms across different modalities into $\mathbf{L}$ as $\alpha_{1} \mathbf{L}_{\text{point}} + \alpha_{2} \mathbf{L}_{\text{p2v}} + \alpha_{3} \mathbf{L}_{\text{point2pixel}} + \alpha_{4} \mathbf{L}_{\text{pixel2point}}$, where $\alpha_{1} \sim \alpha_{4}$ are set as 1.0, 1.0, 0.5, 1.0 to balance loss terms.

\begin{table}[!t]
\caption{Data augmentation transformations on LiDAR (L) point cloud and camera (C) images. We only treat random flipping as a naive symmetric transformation that can be simply applied to both modalities simultaneously.}
\label{tab:lidarDA&cameraDA}
\vspace{-3mm}
\small
\setlength{\tabcolsep}{0.6mm}
\centering
\resizebox{\linewidth}{!}{%
\begin{tabular}{@{}l|l@{}}
\toprule
L-only & \begin{tabular}[c]{@{}l@{}}
$\bullet$ Global rotation around the $Z$ axis with a random angle in $\left[- \frac{\pi}{4}, + \frac{\pi}{4}\right]$.\\ 
$\bullet$ Global translation with a random vector $(\Delta x, \Delta y, \Delta z)$ sampled \\~~from a Gaussian distribution with mean zero and the standard deviation 0.5.\\ 
$\bullet$ Global scaling with a random scaling factor in $\left[0.95, 1.05\right]$.\\
...
\end{tabular} \\ 
\midrule
Symmetric & $\bullet$ Random flipping along the $X, Y$ axis with probability 0.5. \\ 
\midrule
C-only & \begin{tabular}[c]{@{}l@{}}
$\bullet$ Scaling with a random ratio in $\left[1.0, 1.5\right]$.\\ 
$\bullet$ Horizontal rotation with a random angle in $\left[\ang{-1}, \ang{1}\right]$.\\
$\bullet$ Random cropping with a size $(H_{\text{in}}, W_{\text{in}})$.\\ 
$\bullet$ Color jitter from torchvision \cite{torchvision_colorjitter} with random parameters of \\~~0.3 for brightness, 0.3 for contrast, 0.3 for saturation, 0.1 for hue.\\ 
$\bullet$ JPEG with a random compression ratio in $\left[30, 70 \right]$ and probability 0.5 \cite{jpeg_training_cvpr2016}.\\ 
...
\end{tabular} \\ 
\bottomrule
\end{tabular}%
}
\vspace{-4.5mm}
\end{table}

\begin{table*}[!t]
\caption{Performance comparison on nuScenes testing set \cite{nusclidarseg_leaderboard}. The modalities available on nuScenes include LiDAR (L), Camera (C), and Radar (R). Top-1 results are in bold. *Submission entries without a published paper by the CVPR 2023 deadline of Nov 11, 2022.}
\label{tab:miou_testset_nusc}
\vspace{-3mm}
\setlength{\tabcolsep}{0.9mm}
\centering
\resizebox{0.75\linewidth}{!}{%
\begin{tabular}{@{}c|c|cc|cccccccccccccccc@{}}
\toprule
Method & \rotatebox{90}{Modality} & \rotatebox{90}{\textbf{mIoU}} & \rotatebox{90}{\textbf{fwIoU}}& \rotatebox{90}{barrier} & \rotatebox{90}{bicycle} & \rotatebox{90}{bus} & \rotatebox{90}{car} & \rotatebox{90}{cons. vehicle} & \rotatebox{90}{motorcycle} & \rotatebox{90}{pedestrian} & \rotatebox{90}{traffic-cone} & \rotatebox{90}{trailer} & \rotatebox{90}{truck} & \rotatebox{90}{driveable} & \rotatebox{90}{other} & \rotatebox{90}{sidewalk} & \rotatebox{90}{terrain} & \rotatebox{90}{manmade} & \rotatebox{90}{vegetation} \\ 
\hline
PolarNet\cite{PolarNet_cvpr2020} & L & 69.42 & 87.38 & 72.16 & 16.81 & 77.01 & 86.53 & 51.14 & 69.65 & 64.80 & 54.11 & 69.70 & 63.53 & 96.64 & 67.14 & 77.70 & 72.13 & 87.13 & 84.47 \\
JS3C-Net\cite{js3cnet_aaai2022} & L & 73.60 & 88.06 & 80.14 & 26.15 & 87.79 & 84.54 & 55.17 & 72.56 & 71.28 & 66.26 & 76.79 & 71.11 & 96.80 & 64.47 & 76.86 & 74.09 & 87.48 & 86.10 \\
Cylinder3D\cite{cylinder3d_cvpr2021} & L & 77.16 & 89.92 & 82.76 & 29.75 & 84.34 & 89.41 & 63.03 & 79.29 & 77.21 & 73.40 & 84.55 & 69.17 & 97.66 & 70.24 & 80.29 & 75.51 & 90.41 & 87.55 \\
AMVNet\cite{AMVNet_arxiv2020} & L & 77.27 & 90.08 & 80.64 & 31.96 & 81.73 & 88.93 & 67.07 & 84.33 & 76.11 & 73.48 & 84.87 & 67.30 & 97.37 & 67.37 & 79.41 & 75.45 & 91.45 & 88.69 \\
SPVNAS\cite{spvnas_eccv2020} & L & 77.35 & 89.68 & 80.00 & 29.98 & 91.92 & 90.81 & 64.68 & 78.99 & 75.62 & 70.94 & 81.01 & 74.64 & 97.44 & 69.23 & 79.95 & 76.10 & 89.28 & 87.06 \\
Cylinder3D++\cite{cylinder3d_cvpr2021} & L & 77.86 & 89.93 & 82.76 & 33.89 & 84.34 & 89.41 & 69.63 & 79.42 & 77.26 & 73.40 & 84.55 & 69.41 & 97.66 & 70.24 & 80.29 & 75.51 & 90.42 & 87.55 \\
AF2S3Net\cite{AF2-S3Net_cvpr21} & L & 78.34 & 88.51 & 78.87 & 52.21 & 89.93 & 84.17 & \textbf{77.42} & 74,30 & 77.32 & 71.95 & 83.88 & 73.78 & 97.13 & 66.47 & 77.51 & 74.01 & 87.69 & 86.80 \\
SPVCNN++\cite{spvnas_eccv2020} & L & 81.12 & 90.97 & \textbf{86.35} & 43.13 & 91.90 & \textbf{92.18} & 75.90 & 75.72 & 83.44 & 77.31 & 86.82 & \textbf{77.36} & 97.69 & \textbf{71.22} & 81.08 & 77.19 & 91.67 & 88.98 \\
\midrule
LIFusion$^*$\cite{nusclidarseg_leaderboard} & LC & 75.74 & 89.32 & 58.13 & 36.30 & 86.67 & 84.28 & 59.96 & 79.69 & 80.30 & 77.77 & 83.23 & 68.74 & 97.18 & 68.19 & 77.04 & 74.45 & 91.03 & 88.95 \\
PMF\cite{pmf_iccv2021} & LC & 77.03 & 89.34 & 82.11 & 40.33 & 80.94 & 86.42 & 63.72 & 79.22 & 79.75 & 75.86 & 81.17 & 67.05 & 97.28 & 67.69 & 78.05 & 74.48 & 89.94 & 88.46 \\
CPFusion$^*$\cite{nusclidarseg_leaderboard} & LCR & 77.72 & 89.19 & 83.67 & 37.03 & 89.02 & 86.24 & 70.08 & 77.47 & 78.07 & 74.53 & 82.78 & 67.94 & 96.64 & 68.24 & 79.53 & 74.91 & 90.47 & 86.95 \\
2D3DNet\cite{2d3dnet_3dv2021} & LC & 79.96 & 90.08 & 83.01 & \textbf{59.35} & 87.99 & 85.09 & 63.70 & \textbf{84.39} & 81.95 & 75.96 & 84.79 & 71.93 & 96.88 & 67.35 & 79.81 & 75.96 & 92.05 & 89.18 \\
\midrule
MSeg3D & LC & \textbf{81.14} & \textbf{91.35} & 83.11 & 42.46 & \textbf{94.92} & 92.01 & 67.10 & 78.58 & \textbf{85.66} & \textbf{80.47} & \textbf{87.53} & 77.32 & \textbf{97.74} & 69.82 & \textbf{81.22} & \textbf{77.83} & \textbf{92.35} & \textbf{90.07} \\

\bottomrule
\end{tabular}%
}
\vspace{-4.5mm}
\end{table*}

\subsection{Asymmetric Multi-modal Data Augmentation}\label{sec:asym_MDA}
The segmentation output and our multi-modal fusion mechanism are point-centered. In Eq.~\ref{eq:bilinear_interpolate} and Eq.~\ref{eq:y_img}, the point and the pixel always can be bridged by the coordinate pair of $(x_{i}, y_{i}, z_{i})$ and $(c_{i}, u_{i}, v_{i})$, as long as we pre-compute the coordinate pairs and keep the order of coordinate pairs among all the points in synchronization. With this ordering constraint, we have the flexibility to decouple multi-modal data augmentation:
\textbf{i}) We can preserve all the LiDAR transformations in Tab.~\ref{tab:lidarDA&cameraDA} with the images and the coordinates $(c,u,v)$ unchanged.
\textbf{ii}) Once coordinates $(c,u,v)$ are synchronously transformed with images, we can further apply the camera-only transformations in Tab.~\ref{tab:lidarDA&cameraDA} to images, which are asymmetric to the LiDAR transformations.
\textbf{iii}) We can apply independent transformations between the images of local cameras since the image view is naturally sliced by local cameras.

The asymmetry lies in not only the LiDAR and camera world but also the local cameras. Thus, we significantly diversify the multi-modal data augmentation for 3D semantic segmentation using the introduced camera transformations with the preserved LiDAR transformations. More potential transformations can be exploited beyond Tab.~\ref{tab:lidarDA&cameraDA}.

\begin{table}[!t]
\caption{Performance comparison on Waymo testing set\cite{waymo3dseg_leaderboard}. *Submission entries without a published paper by the CVPR 2023 deadline of Nov 11, 2022.}
\label{tab:miou_testset_waymo}
\vspace{-3mm}
\centering
\resizebox{0.4\linewidth}{!}{%
\begin{tabular}{@{}c|c@{}}
\toprule
Method & \textbf{mIoU} \\ \midrule
LiDARMultiNet$^*$\cite{LidarMultiNet_arxiv2022} & 71.13 \\
HorizonSegExpert$^*$\cite{waymo3dseg_leaderboard} & 69.44 \\
SPVCNN++\cite{spvnas_eccv2020} & 67.70 \\
PolarFuse$^*$\cite{waymo3dseg_leaderboard} & 67.28 \\
SalsaNext\cite{salsanext_springer2020} & 55.85 \\
\midrule
MSeg3D & 70.51 \\ \bottomrule
\end{tabular}}
\vspace{-2mm}
\end{table}

\section{Experiment}
\subsection{Experimental Setup}

\textbf{nuScenes Dataset} splits 28,130 training samples, 6,019 validation samples, and 6,008 testing samples \cite{nuscens_dataset,PanopticNuscenes_dataset}. Each nuScenes sample contains relatively sparser point cloud of 32 beams and RGB images captured by 6 cameras: front, front-left, front-right, back, back-left, and back-right. Points outside are projected below the image bottom due to different vertical FOVs of LiDAR and cameras. Following official protocol, the $N_{\text{cls}}$ is 17. Semantic segmentation annotations are only on point clouds and not on images.

\textbf{Waymo Dataset} for 3D semantic segmentation includes 23,691 training samples, 5,976 validation samples, and 2,982 testing samples \cite{waymo_cvpr2020}. Each sample contains the point cloud of 64 beams and the RGB images captured by 5 cameras: front, front-left, front-right, side-left, and side-right. There is no rear camera on Waymo ego-vehicle, hence more points outside unfairly increase the difficulty of multi-modal segmentation. The $N_{\text{cls}}$ is 23. Similarly, semantic segmentation annotations are only on point clouds.

\textbf{SemanticKITTI Dataset} is collected by a LiDAR with 64 beams \cite{semKITTI_dataset}. Following \cite{pmf_iccv2021,cylinder3d_cvpr2021,salsanext_arxiv2020,rpvnet_iccv2021}, we use sequences 00 to 10 (excluding 08) with 19,130 training samples and sequence 08 with 4,071 validation samples. It provides only the images of the front-view camera. The $N_{\text{cls}}$ is 20.

\textbf{Evaluation Metric} is the mean Intersection-over-Union (mIoU) defined as $
    \frac{1}{C} \sum_{c=1}^{N_{\text{cls}}- 1} \frac{ {TP}_{c}}{ {TP}_{c} + {FP}_{c} + {FN}_{c}},$
where ${TP}_{c}$, ${FP}_c$, ${FN}_c$ denote the number of true positive, false positive, and false negative predictions for the $c$-th category out of $N_{\text{cls}} - 1$ valid categories. 
Besides, nuScenes computes another metric fwIoU by weighting category-wise IoU with the point frequency of its category.

\textbf{Settings}.
Although lightweight LiDRA-only baseline models can be trained with larger batchsize and more epochs to obtain relatively better results, we still train all the models under the same schedule: a batch of 32 random samples distributed on 16 Tesla V100 GPUs with 24 epochs.
More implementation details on point cloud voxelization, network architecture, model training, and inference are all included in SM.

\subsection{Comparison with State-of-the-art Methods}

\textbf{Results on nuScenes}.
From Tab.~\ref{tab:miou_testset_nusc}, multi-modal methods counterintuitively lag far behind LiDAR-only methods. In the top-20 submissions on nuScenes benchmark \cite{nusclidarseg_leaderboard}, there are only four multi-modal methods. Details of Cylinder3D++, SPVCNN++, LIFusion, and CPFusion are not available till submission. Our framework achieves the best mIoU and fwIoU with multi-modal fusion mechanism and asymmetric multi-modal data augmentation.
Especially, our method achieves superior performance on challenging small objects such as pedestrians and traffic cones, where the laser points are typically insufficient for LiDAR-only methods. 
Our method also significantly outperforms all public and private multi-modal submissions in Tab.~\ref{tab:miou_testset_nusc}.

\textbf{Results on Waymo}.
We perform evaluation on Waymo benchmark \cite{waymo3dseg_leaderboard} in Tab.~\ref{tab:miou_testset_waymo}, where the LiDAR sensor provides denser laser points and the multi-camera excludes the rearview. Such denser point cloud and incomplete camera FOV unfairly weaken the advantages of multiple modalities. Although, our method still achieves state-of-the-art performance using a general point cloud backbone network \cite{openpcdet2022,parta2_tpami2021}. We believe that the performance can be improved by incorporating stronger point cloud backbone networks from other state-of-the-art LiDAR-only methods.

\textbf{Results on SemanticKITTI}.
While comparisons on the highly competitive nuScenes and Waymo datasets validate the effectiveness of our method, Tab.~\ref{tab:semanticKITTI_val} shows the experimental comparison on SemanticKITTI \cite{semKITTI_dataset}. Since SemanticKITTI provides only front-view images, PMF \cite{pmf_iccv2021} evaluates the multi-modal segmentation performance on points inside the FOV intersection. We follow PMF to report mIoU$\textsuperscript{\ref{footnote:miou1}}$ on points inside in Tab.~\ref{tab:semanticKITTI_val}. Our method not only shows improvements to the LiDAR-only methods but also outperforms other multi-modal methods.

\begin{table}[!t]
\caption{Performance comparison on SemanticKITTI \cite{semKITTI_dataset} validation set following PMF \cite{pmf_iccv2021}. The results of the other methods are from PMF paper.}
\label{tab:semanticKITTI_val}
\vspace{-3mm}
\centering
\resizebox{0.45\linewidth}{!}{%
\begin{tabular}{@{}c|c|c@{}}
\toprule
Method & Modality & \textbf{mIoU}$\textsuperscript{\ref{footnote:miou1}}$ \\ 
\midrule
SalsaNext\cite{salsanext_arxiv2020} & L & 59.4 \\
SPVNAS\cite{spvnas_eccv2020} & L & 62.3 \\
Cylinder3D\cite{cylinder3d_cvpr2021} & L & 64.9 \\
PointPainting\cite{pointpainting_cvpr2020} & LC & 54.5 \\
RGBAL\cite{RGBLIDAR_ITSC2019} & LC & 56.2 \\
PMF\cite{pmf_iccv2021} & LC & 63.9 \\ 
\midrule
L-Baseline & L & 64.8 \\
MSeg3D & LC & 66.7 \\ 
\bottomrule
\end{tabular}%
}
\vspace{-2mm}
\end{table}

\begin{table}[!t]
\caption{Analysis on the performance gap on all points and points inside, and the data augmentation (DA). Only GF-Phase is simply used for multi-modal fusion (M-Fusion) here. DA includes LiDAR DA (L-DA) and Multi-modal DA (M-DA).}
\label{tab:ablation_sinle&multi-modal_nusc&waymo}
\vspace{-3mm}
\centering
\resizebox{0.7\linewidth}{!}{%
\begin{tabular}{@{}c|c|c|cc|cc@{}}
\toprule
\multirow{2}{*}{LiDAR} & \multirow{2}{*}{M-Fusion} & \multirow{2}{*}{DA} & \multicolumn{2}{c|}{nuScenes} & \multicolumn{2}{c}{Waymo} \\ \cmidrule(l){4-7} 
 &  &  & mIoU & mIoU$\textsuperscript{\ref{footnote:miou1}}$ & mIoU & mIoU$\textsuperscript{\ref{footnote:miou1}}$ \\
\hline
$\surd$ & $\times$ & L-DA & 72.00 & 70.76 & 67.48 & 67.41 \\
\hline
$\surd$ & GF-Phase & $\times$ & 68.10 & 76.68 & 59.77 & 64.79 \\
$\surd$ & GF-Phase & L-DA & 71.35 & 77.48 & 60.34 & 65.70 \\
\cellcolor{lightgray}$\surd$ & \cellcolor{lightgray}GF-Phase & \cellcolor{lightgray}M-DA & \cellcolor{lightgray}72.39 & \cellcolor{lightgray}78.65 & \cellcolor{lightgray}63.97 & \cellcolor{lightgray}67.94 \\ \bottomrule
\end{tabular}}
\vspace{-2mm}
\end{table}

\begin{table}[!t]
\caption{Analysis on multi-modal feature fusion of GF-Phase, cross-modal (CM) feature completion, and SF-Phase. The supervised CM feature completion is decomposed as $\mathbf{L}_{\text{pixel2point}}$ (Eq.~\ref{eq:loss_pixel2point}) and ``Comp.'' (Eq.~\ref{eq:mix_pcam_cam}). The gap can be formulated as mIoU - mIoU$\textsuperscript{\ref{footnote:miou1}}$.}
\label{tab:effect_of_fusion_phases}
\vspace{-3mm}
\small
\setlength{\tabcolsep}{0.6mm}
\centering
\resizebox{\linewidth}{!}{%
\begin{tabular}{@{}c|cc|cc|ccc|ccc@{}}
\toprule
\multicolumn{1}{c|}{\multirow{2}{*}{LiDAR}} & \multicolumn{2}{c|}{M-Fusion} & \multicolumn{2}{c|}{CM Feature Completion}  & \multicolumn{3}{c|}{nuScenes} & \multicolumn{3}{c}{Waymo} \\ \cmidrule(lr){2-11}
\multicolumn{1}{c|}{} & \multicolumn{1}{c}{GF-Phase} & \multicolumn{1}{c|}{SF-Phase} & \multicolumn{1}{c}{$\mathbf{L}_{\text{pixel2point}}$} & \multicolumn{1}{c|}{Comp.} & \multicolumn{1}{c}{mIoU} & \multicolumn{1}{c}{mIoU$\textsuperscript{\ref{footnote:miou1}}$} & \multicolumn{1}{c|}{Gap} & \multicolumn{1}{c}{mIoU} & \multicolumn{1}{c}{mIoU$\textsuperscript{\ref{footnote:miou1}}$} & \multicolumn{1}{c}{Gap} \\ 
\hline
$\surd$ & $\times$ & $\times$ & $\times$ & $\times$ & 72.00 & 70.76 & +1.24 & 67.48 & 67.41 & +0.07  \\
\hline
$\surd$ & $\surd$ & $\times$ & $\times$ & $\times$ & 72.39 & 78.65 & -6.26 & 63.97 & 67.94 & -3.97 \\ 
$\surd$ & $\surd$ & $\times$ & $\surd$ & $\times$ & 76.44 & 79.10 & -2.66 & 67.13 & 69.06 & -1.93 \\
$\surd$ & $\surd$ & $\times$ & $\surd$ & $\surd$ & 78.28 & 79.10 & -0.82 & 67.89 & 69.06 & -1.17 \\
\cellcolor{lightgray}$\surd$ & \cellcolor{lightgray}$\surd$ & \cellcolor{lightgray}$\surd$ & \cellcolor{lightgray}$\surd$ & \cellcolor{lightgray}$\surd$ & \cellcolor{lightgray}80.00 &\cellcolor{lightgray} 80.10 & \cellcolor{lightgray}-0.10 & \cellcolor{lightgray}69.63 & \cellcolor{lightgray}70.19 & \cellcolor{lightgray}-0.56 \\ 
\bottomrule
\end{tabular}%
}
\vspace{-4.5mm}
\end{table}

\subsection{Ablation Study}\label{sec:ablation_study}
\vspace{-2mm}
\textbf{Difficulties Arising from Multi-modality} are investigated in Tab.~\ref{tab:ablation_sinle&multi-modal_nusc&waymo} from two perspectives of the performance gap between all points and points inside, as well as the multi-modal data augmentation. 
The experiments in Tab.~\ref{tab:ablation_sinle&multi-modal_nusc&waymo} start with a vanilla variant of the multi-modal model using the basic GF-Phase fusion without the multi-modal data augmentation.
The mIoU is relatively worse than the mIoU$\textsuperscript{\ref{footnote:miou1}}$, since the geometry-based feature fusion ideally ignores the inevitable points outside with the missing camera features \cite{pmf_iccv2021, RGBLIDAR_ITSC2019,fuseseg_wacv2020}, which also motivates us to close the performance gaps between mIoU and mIoU$\textsuperscript{\ref{footnote:miou1}}$ by using the cross-modal feature completion and semantic-based SF-Phase.
Besides, in rows 2 and 1, the multi-modal segmentation performance will be worsened if we deprecated the augmentation transformations that cannot be directly applied to both modalities. However, the proposed asymmetric multi-modal data augmentation allows the accumulation of the transformations on point cloud and image, achieving the best performance on both datasets.

\textbf{Multi-modal Feature Fusion Modules}.
Tab.~\ref{tab:effect_of_fusion_phases} shows that our supervised cross-modal feature completion and SF-Phase benefit both mIoU and mIoU$\textsuperscript{\ref{footnote:miou1}}$ with gradually narrowed gaps.
In row 3, $\mathbf{L}_{\text{pixel2point}}$ can also bring improvements, which indicates that transferring appearance information from dense image to sparse point cloud is applicable and beneficial for joint feature learning. In row 4, completing camera features with pseudo-camera features facilitates the feature gap reduction between points inside and outside. Eventually, the SF-Phase further narrows the gaps between mIoU and mIoU$\textsuperscript{\ref{footnote:miou1}}$ to the smallest values of 0.10 and 0.56, addressing the limitation of geometric associations by semantic-based feature fusion. Thus, we set the model in the last row as our final framework.

\begin{figure}[!t]
	\centering
	\includegraphics[width=0.95\linewidth]{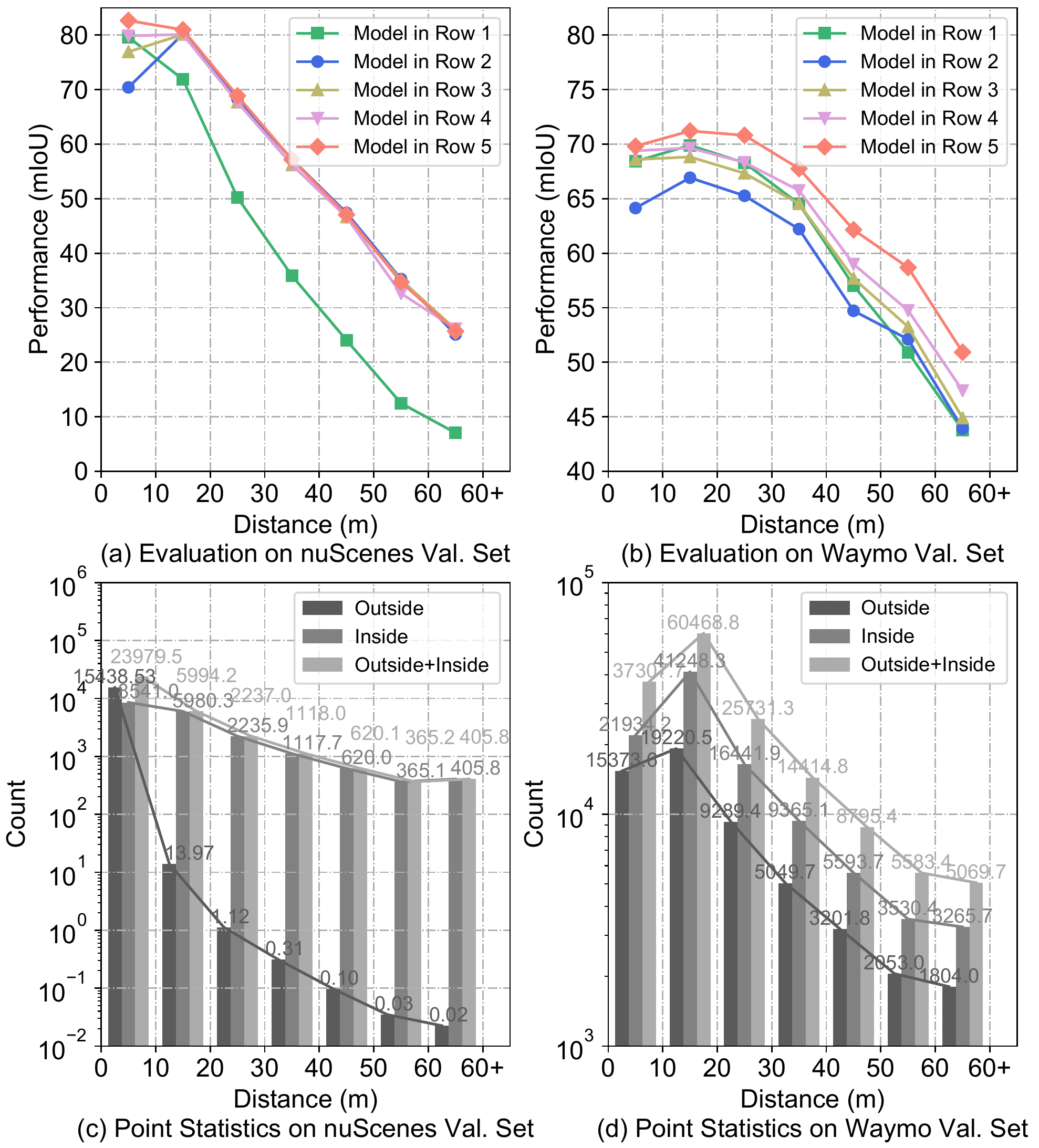}%
	\vspace{-3mm}
 	\caption{Distance-based mIoU evaluation on nuScenes and Waymo validation (Val.) sets using the models in Tab.~\ref{tab:effect_of_fusion_phases}.}
	\label{fig:distance_miou_nusc_waymo_2x2}
	\vspace{-3mm}
\end{figure}

\textbf{mIoU Breakdown Over Distance}.
Fig.~\ref{fig:distance_miou_nusc_waymo_2x2} presents the distance-based evaluation corresponding to the models in Tab.~\ref{tab:effect_of_fusion_phases}. The LiDAR-only model degrades at long distances due to more sparse points. Instead, the multi-modal model in row 5 effectively alleviates the performance degradation.
In Fig.~\ref{fig:distance_miou_nusc_waymo_2x2} (a) and (b), improvements between the multi-modal model in row 5 and the LiDAR-only model in row 1 are also increased along the distance due to increased sparsity. 
The reasons why the improvement of the multi-modal model on Waymo is not as similar as that on nuScenes are further analyzed in Fig.~\ref{fig:distance_miou_nusc_waymo_2x2} (c) and (d).
The Waymo point cloud has denser points, which reduces the dependence on the image. More points outside are distributed along different distances due to no rear cameras on Waymo.
Although more points outside hinder the applicability of multi-modal fusion, our final multi-modal model still yields a low-performance gap of 0.56 mIoU on Waymo in row 5.

\begin{table}[!t]
\caption{Robustness analysis on MSeg3D by removing some cameras as malfunction. ``\#Camera'' denotes the number of available cameras. ``$\times$'' denotes LiDAR-only baseline.}
\label{tab:ablation_cam_malfunctions}
\vspace{-4mm}
\setlength{\tabcolsep}{0.8mm}
\centering
\resizebox{0.8\linewidth}{!}{%
\begin{tabular}{@{}c|ccccccc|c@{}}
\toprule
\#Camera & 6 & 5 & 4 & 3 & 2 & 1 & 0 & $\times$ \\ \midrule
nuScenes & 80.00 & 78.69 & 77.62 & 76.69  & 76.01  & 75.44 & 74.47 & 72.00 \\
Waymo & - & 69.63 & 69.18 & 68.86 & 68.13 & 68.09 & 68.04  & 67.48 \\ \bottomrule
\end{tabular}}
\vspace{-2mm}
\end{table}

\begin{table}[!t]
\caption{Robustness analysis on MSeg3D with multi-frame point clouds input. ``\#L-Frame'' is the frame number for LiDAR.
}
\label{tab:ablation_multiframe_pc_input}
\vspace{-3.5mm}
\setlength{\tabcolsep}{0.8mm}
\centering
\resizebox{0.8\linewidth}{!}{%
\begin{tabular}{@{}c|c|cccccc|c@{}}
\toprule
\multirow{3}{*}{nuScenes} & \#L-Frame & 1 & 10 & 20 & \cellcolor{lightgray}25 & 30 & 40 & 25 \\ 
 & Camera & $\times$ & $\times$ & $\times$ & \cellcolor{lightgray}$\times$ & $\times$ & $\times$ & $\surd$ \\
 & mIoU & 72.00 & 74.66 & 75.37 & \cellcolor{lightgray}75.77 & 75.28 & 75.15 & 81.12 \\ \midrule
\multirow{3}{*}{Waymo} & \#L-Frame & 1 & 5 & \cellcolor{lightgray}10 & 15 & 20 & 30 & 10 \\
 & Camera & $\times$ & $\times$ & \cellcolor{lightgray}$\times$ & $\times$ & $\times$ & $\times$ & $\surd$ \\
 & mIoU & 67.48 & 69.18 & \cellcolor{lightgray}69.45 & 68.88 & 68.78 & 68.64 & 70.20 \\ \bottomrule
\end{tabular}}
\vspace{-5mm}
\end{table}

\textbf{Robustness Against Camera Malfunction}.
In Tab.~\ref{tab:ablation_cam_malfunctions}, our MSeg3D performs properly under the unfavorable condition of camera malfunction, and it can easily deal with the practical situation without switching the model. Note that even with all cameras removed, MSeg3D still outperforms the LiDAR-only baseline, where our cross-modal feature completion supervision provides effective cross-modal information transfer in training.

\textbf{Robustness Against Multi-frame Point Clouds Input}.
Since the densified points alleviate the sparsity, collapsing laser points from neighboring frames usually boosts LiDAR-only 3D perception \cite{js3cnet_aaai2022,WYSISWYG_CVPR2020}. 
In Tab.~\ref{tab:ablation_multiframe_pc_input}, we follow \cite{centerpoint_cvpr2021} collapsing multiple previous frames to the current frame from the provided ego-vehicle motion information. For LiDAR-only model, the improvements are saturated with 25 and 10 frames on nuScenes and Waymo. Under such conditions, our MSeg3D still achieves further improvements in the last column, which shows that multi-frame point clouds input can be an optional extension.

\textbf{Complexity Scalability}.
From Tab.~\ref{tab:ablation_scalability_nusc}, a lightweight image backbone also delivers significant performance gains due to multi-modal input.
The best-performing HRNet-w48 is our default image backbone.
However, the multi-camera image input makes the efficiency bottleneck lie in the image backbone, which deserves in-depth study for autonomous driving \cite{bevformer_eccv2022}.
We believe that real-time image segmentation networks such as BiSeNet \cite{BiSeNet_eccv2018,BiSeNet2_ijcv2021} and Fast-SCNN \cite{FastSCNN_BMVC2019} can accelerate our approach.
Overall, our MSeg3D can adapt to complexity scalability among various backbones, which flexibly trades off performance and efficiency.

\begin{table}[!t]
\caption{Scalability analysis by barely varying the backbone settings on nuScenes validation set.}
\label{tab:ablation_scalability_nusc}
\vspace{-3mm}
\setlength{\tabcolsep}{0.6mm}
\centering
\resizebox{0.6\linewidth}{!}{%
\begin{tabular}{@{}c|c|c|c@{}}
\toprule
Image Backbone & mIoU & \#Params(M) & Latency(s) \\ \midrule
$\times$ & 72.00 & 21.28  & 0.083 \\
SegFormer-B0\cite{segformer_nips2021} & 78.14  & 25.33  & 0.204 \\
SegFormer-B5\cite{segformer_nips2021} & 78.89  & 103.48  & 0.479 \\
HRNet-w18\cite{hrnet_tpami2021} & 79.21  & 31.56  & 0.265 \\
ResNet101\cite{resnet_cvpr2016} & 79.36  & 64.60  & 0.567 \\
\cellcolor{lightgray}\cellcolor{lightgray}HRNet-w48\cite{hrnet_tpami2021} & \cellcolor{lightgray}80.00 & \cellcolor{lightgray}87.34  & \cellcolor{lightgray}0.445 \\ 
\bottomrule
\end{tabular}%
}
\vspace{-5mm}
\end{table}

\vspace{-2mm}
\section{Conclusion}\label{sec:conclusion}
\vspace{-1.5mm}
We propose a novel multi-modal 3D semantic segmentation method termed MSeg3D for autonomous driving, based on LiDAR and multi-camera sensors.
Our cross-modal feature completion and semantic-based feature fusion solve the overlooked problem, where multi-modal fusion can occur only in the sensor FOV intersection.
The proposed asymmetric multi-modal data augmentation enables the multi-modal segmentation model to be effectively trained with reliable performance.
Our method achieves state-of-the-art performance on nuScenes, Waymo, and SemanticKITTI. The comprehensive experiments validate improvements and robustness. 
We hope that our work can inspire further investigation into multi-modal fusion for 3D semantic segmentation in autonomous driving.

\textbf{Acknowledgment}: This work was supported by the National Key Research and Development Program of China (2018YFE0183900) and YUNJI Technology Co. Ltd.


\end{document}